%% file: main.tex
\documentclass[10pt,twocolumn,letterpaper]{article}
\usepackage[final]{cvpr} 

\usepackage{amsmath}
\usepackage{graphicx}
\usepackage{textcomp}

\usepackage[detect-none,binary-units=true]{siunitx}
\DeclareSIUnit{\nothing}{\relax}
\usepackage{arydshln}
\usepackage{multirow}
\usepackage{booktabs}
\usepackage{gensymb}
\usepackage[dvipsnames]{xcolor}
\usepackage{algorithmic}
\usepackage{algorithm2e}
\usepackage[para,online,flushleft]{threeparttable}

\definecolor{cvprblue}{rgb}{0.21,0.49,0.74}
\usepackage[pagebackref,breaklinks,colorlinks,citecolor=cvprblue]{hyperref}

\def\paperID{15} 
\def\confName{CVPR}
\def\confYear{2024}

\title{Multi-resolution Rescored ByteTrack for Video Object Detection \\ on Ultra-low-power Embedded Systems}
\author{
    *Luca Bompani$^1$
    \quad
     Manuele Rusci$^2$
    \quad
    Daniele Palossi$^{3,4}$ 
    \quad
    Francesco Conti$^1$ 
    \quad
    Luca Benini$^{1,4}$\\
    $^1$ Department of Electrical, Electronic and Information Engineering, University of Bologna, Italy \\ $^2$ Department of Electrical Engineering, KU Leuven, Belgium \\ $^3$Dalle Molle Institute for Artificial Intelligence, USI-SUPSI, Switzerland \\ $^4$Integrated Systems Laboratory, ETH Z\"urich, Switzerland \\
    *{\tt\small luca.bompani5@unibo.it}
}

\begin{document}
\maketitle


\input{01-abstract}
\input{02-introduction}
\input{03-related-works}
\input{04-methodology}
\input{05-GAP9}
\input{06-results}
\input{07-conclusions}

\section*{Acknowledgements}
This work has been partially supported by the Swiss SNF project 207913 TinyTrainer: On-chip Training for TinyML devices.

{\small 
\bibliographystyle{ieeenat_fullname}
\bibliography{main}
}

\end{document}

%% file: 01-abstract.tex
\begin{abstract}
This paper introduces Multi-Resolution Rescored ByteTrack (MR2-ByteTrack), a novel video object detection framework for ultra-low-power embedded processors. 
This method reduces the average compute load of an off-the-shelf Deep Neural Network (DNN) based object detector by up to 2.25$\times$ by alternating the processing of high-resolution images (320$\times$320 pixels) with multiple down-sized frames (192$\times$192 pixels). 
To tackle the accuracy degradation due to the reduced image input size, MR2-ByteTrack correlates the output detections over time using the ByteTrack tracker and corrects potential misclassification using a novel probabilistic Rescore algorithm. 
By interleaving two down-sized images for every high-resolution one as the input of different state-of-the-art DNN object detectors with our MR2-ByteTrack, we demonstrate an average accuracy increase of 2.16\% and a latency reduction of 43\% on the GAP9 microcontroller compared to a baseline frame-by-frame inference scheme using exclusively full-resolution images. 
Code available at: \url{https://github.com/Bomps4/Multi_Resolution_Rescored_ByteTrack}
\end{abstract}

%% file: 02-introduction.tex
\section{Introduction} \label{sec:introduction}

A Video Object Detection (VOD) algorithm identifies objects in a video stream by reporting their categories and the corresponding coordinates in the image space.
This function is critical for various cyber-physical systems like surveillance cameras~\cite{ULP_Camera} and miniaturized (as big as the palm of a hand) drones~\cite{lamberti2023bio,alnuaimi2022deep}. 
These systems typically integrate a camera with an ultra-low-power embedded processor, i.e., a low-cost Microcontroller Unit (MCU). 
Compared to other edge/mobile processors (e.g., Nvidia Tegra, Raspberry Pi, Qualcomm Snapdragon, etc.), MCUs feature a 10-100$\times$ power consumption to enable battery-powered operations (also referred to as the \textit{extreme edge computing}~\cite{Extreme_edge}).  
On the other side, these devices present a limited on-chip memory (typically a few \SI{}{\mega\byte}) and low computational power~\cite{VegaDSP}.
These severe limitations challenge the design of high-throughput VOD systems using compute-intensive accurate vision algorithms.

\begin{figure}[t]
\includegraphics[width=\columnwidth]{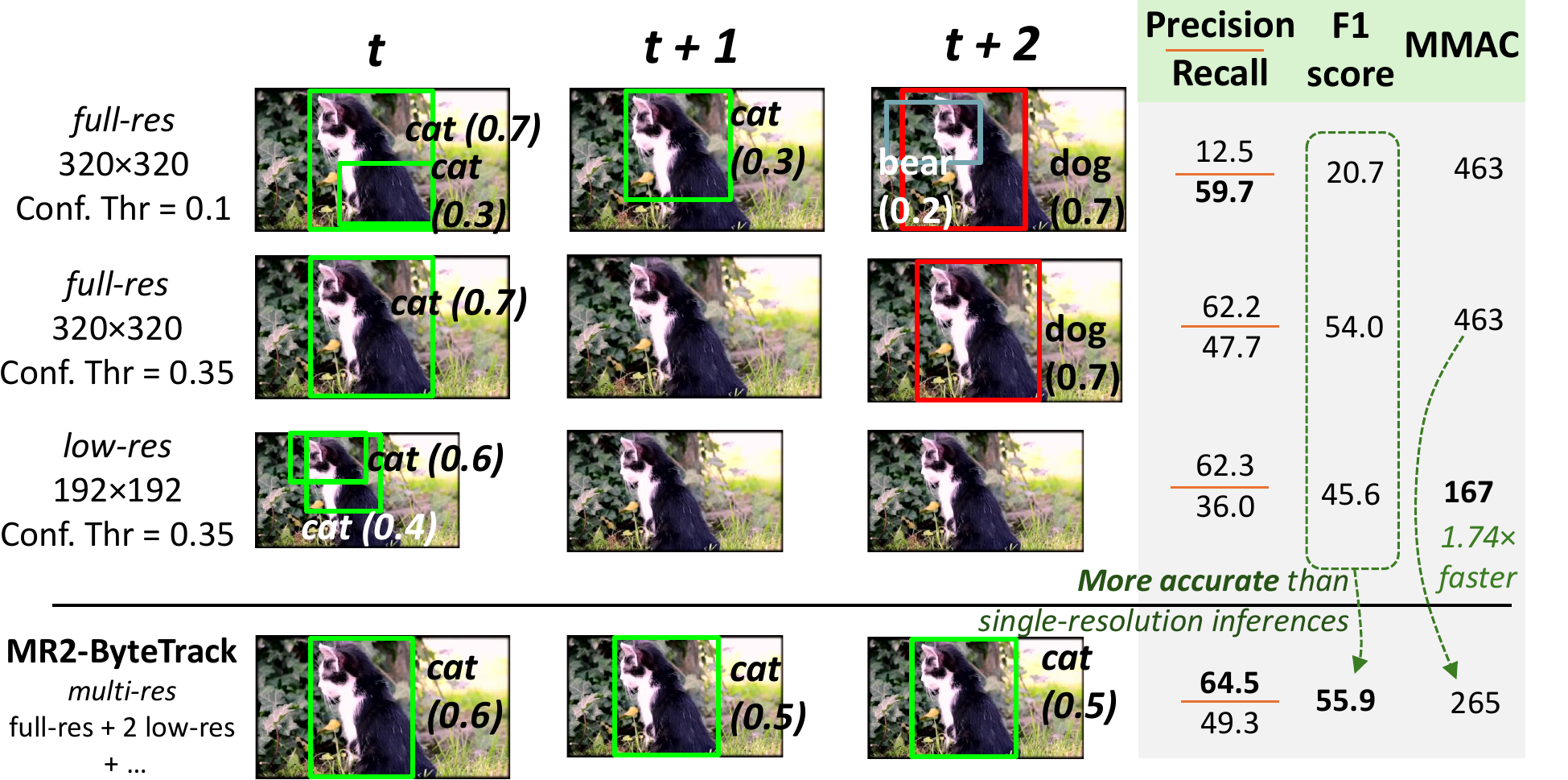}
\caption{Video object detection using NanoDet-Plus~\cite{nanodet} object detector under multiple thresholds and input size settings vs. the proposed \textit{MR2-ByteTrack} solution.}
\label{fig:motivation}
\end{figure}

Recently, VOD was shown on multi-core MCUs~\cite{VegaDSP} using lightweight (i.e., with a low memory budget) Deep Neural Network-based (DNN) object detectors with a simple frame-by-frame analysis~\cite{lamberti2023bio,lamberti2021low,alnuaimi2022deep}.
For every frame of the image stream, the DNN inference task returns a list of output detections with confidence scores between 0 and 1. 
For example, Fig.~\ref{fig:motivation} shows the output of the recent NanoDet-Plus object detector~\cite{nanodet} featuring 1.17 million parameters when applied over three consecutive video frames.
The precision-recall tradeoff is adjusted by thresholding the confidence scores, as shown in the figure: a higher threshold value leads to fewer false positives (higher precision) or undesired miss-detections (lower recall).
From a computational cost perspective, a common strategy to reduce the number of operations per frame, and thus increasing the processing throughput, consists of feeding the model with low-resolution (low-res) input images~\cite{redmon2018yolov3,DynamicRes}, e.g.,~-\SI{296}{\mega\nothing} Multiply-Accumulate (MAC) using \SI{2.8}{\nothing}$\times$ smaller images in our example. 

To tackle this problem, this paper proposes \textit{Multi-Resolution Rescored ByteTrack} (\textit{MR2-ByteTrack}), a method that transforms an image object detection model into a full VOD algorithmic pipeline, simultaneously improving the detection accuracy and reducing the overall computational cost with respect to a naïve \textit{frame-by-frame} processing pipeline. 
MR2-ByteTrack combines an off-the-shelf DNN object detector with an extended version of \textit{ByteTrack}~\cite{ByteTrackMT}. 
This lightweight Kalman-filter-based tracker updates the output detections based on the object instances already detected by the DNN. 
Because the original ByteTrack~\cite{ByteTrackMT} lacks recovery mechanisms from misclassification (e.g., a cat is tracked as a ``dog'' if wrongly predicted at the beginning of the video sequence), we propose a novel \textit{Rescore} algorithm to refine classifications over time, according to a probabilistic logic. 

Our approach adopts a \textit{Multi Resolution Inference} scheme to reduce the per-frame workload, inserting low-resolution (low-res) frames between full-resolution (full-res).
As shown at the bottom of Fig.~\ref{fig:motivation}, our MR2-ByteTrack leverages the temporal correlation of detections in consecutive frames to recover the accuracy drop of low-resolution inferences.
When deployed on a multi-core MCU, this method incurs no additional memory cost for parameter storage compared to a single-resolution inference scheme, as the same object detection DNN is applied to both full and low-resolution frames.

\textbf{In summary, this work makes the following novel contributions:}
\begin{itemize}
\item The MR2-ByteTrack framework, combining an off-the-shelf DNN-based object detector for multi-resolution inference, the ByteTrack Kalman-based tracker, and the \textit{Rescore} method to refine category assignment of tracked frames, reducing misdetections and misclassifications;
\item The first embodiment of a multi-resolution VOD pipeline on an ultra-low-power MCU, Greenwaves Technologies (GWT)'s GAP9, showing a better accuracy-throughput tradeoff than single-resolution inference VOD systems.
\end{itemize}

We validate our proposed MR2-ByteTrack method using three State-of-the-Art (SoA) tiny object detectors: YOLOX-Nano~\cite{ge2021yolox}, NanoDet-Plus~\cite{nanodet}, and EfficientDet-D0~\cite{EfficientDET}, all off-the-shelf pretrained on the COCO dataset~\cite{cocodataset}. 
Across the ImageNetVid~\cite{ILSVRC15} validation set, MR2-ByteTrack improves mean Average Precision (mAP) scores compared to the object detector baselines by up to 5.17\% and the F1 score by 3.58\% if applied only on full-res frames. 
If, instead, we interleave for each full-res frame two low-res frames, the mAP improvement is +2.16\%  while the average MAC computational costs are reduced by up to 43\%. 

We deployed our multi-resolution VOD strategy, using the on the GAP9 MCU processor featuring a 9-core compute cluster at a peak power cost of up to \SI{72}{\milli\watt}, also considering the external \SI{32}{\mega\byte} and \SI{8}{\mega\byte} MB FLASH and RAM.  
Compared to the baseline object detectors~\cite{ge2021yolox,nanodet,EfficientDET}, MR2-ByteTrack achieves up to 1.76$\times$ lower inference latency with no increase in the DNN’s parameters footprint (\SI{2.24}{\mega\byte} for the largest model) and only a modest increase in code size (+\SI{186}{\kilo\byte}), thus enabling high-accuracy VOD on milliwatt-power extreme-edge devices.
Also, when compared to the SoA YOLOV method~\cite{2023yolov}, our MR2-ByteTrack shows the same F1 score but saves the memory and computes overheads (14\% and 21\%) of the transformer-based front-end that differently, from our ByteTrack module, demands an ad-hoc training process. 

%% file: 03-related-works.tex
\section{Related Work} \label{sec:related-works}

\subsection{Lightweight Object Detection} 
DNN-based object detection algorithms can be grouped into two main categories: \textit{two-stage} and \textit{one-stage} detectors. 
The two-stage methods, e.g., Faster R-CNN~\cite{renNIPS15fasterrcnn}, generate a set of region proposals that are analyzed during a second inference stage.
This dynamic and unpredictable workload, as it depends on the generated proposals, makes these methods unsuitable for resource-constrained embedded devices, where low computation and predictable execution time are crucial to optimally distribute the limited resources.
Consequently, many recent works focus on one-stage memory-efficient DNNs, generally composed of a feature extractor and multiple head blocks that predict the bounding box coordinates~\cite{MobileNetv2, nanodet, Xiong_2021_CVPR, ge2021yolox, EfficientDET}.

SSDLite~\cite{MobileNetv2} adds detector heads on top of a MobileNetV2 backbone and achieves an mAP of \SI{22.1}{\nothing} on the COCO dataset~\cite{cocodataset} with a total memory and computational cost of \SI{4.3}{\mega\nothing} parameters and \SI{0.8}{\giga MAC}, respectively.
An instance of this model has been successfully demonstrated on an MCU device with a peak power of \SI{150}{\milli\watt} for object detection aboard nano-drones~\cite{lamberti2023bio} or license plate detection for smart cameras~\cite{lamberti2021low}, showing a peak throughput of $\sim$\SI{1}{\nothing} Frame Per Second (FPS).
NanoDet-Plus~\cite{nanodet} introduces an ``assign guidance module'' and a dynamic soft label assigner to improve performance, reaching a mAP of \SI{27}{\nothing} on the COCO dataset with \SI{0.45}{\giga MAC} operations and \SI{1.17}{\mega\nothing} parameters.

YOLOX-Nano~\cite{ge2021yolox} starts from the YOLOv3 architecture and introduces depthwise convolutions and an anchor-free decoupled head, which lead to a mAP of \SI{25.4}{\nothing} on COCO with \SI{0.8}{\giga MAC} and \SI{0.9}{\mega\nothing} parameters. 
Instead, EfficientDet-D0~\cite{EfficientDET}, leveraging a neural architecture search technique, introduces a bidirectional feature pyramid network to aggregate features that, in conjunction with a decoupled head, pushes the mAP to \SI{34.6}{\nothing} at the cost of \SI{2.5}{\giga MAC} and \SI{3.9}{\mega\nothing} parameters.
Our work builds on these recent frame-by-frame models to construct a novel framework operating on video streams. 

\subsection{Video Object Detection} 
A first class of techniques for video object detection uses 3D convolutions instead of 2D ones~\cite{tran2015learning}. 
This conceptually simple approach is not real-time and demands a large memory footprint to store high-dimensional activation tensors, exceeding the memory available on low-end MCU systems. 
An alternative line of research aggregates information across frames from region proposals obtained by two-stage detectors: in~\cite{FGFA} using optical flow, or convolutional-based trackers in~\cite{LYU2021139, IntegratedOD}, and transformer-based memory layers in~\cite{MEGA}. 
However, the cost of region proposal aggregation exceeds the available budget for MCUs. 

A more lightweight group of techniques builds on top of one-stage detectors by modeling the temporal correlation of detections.
\cite{seqnms} uses dynamic programming to link objects across frames based on the Intersection over Union (IoU) metric. 
\textit{Seq-Bbox}~\cite{Seq-BBox} and \textit{Motion-based Seq-Bbox}~\cite{motion_based} refine the outputs of the underlying object detector using the temporally associated detection information, marking a mAP of \SI{80.9}{\nothing} and \SI{72.7}{\nothing}, respectively, on the ImagenetVID dataset (+\SI{6.9}{\nothing} and +\SI{5.5}{\nothing} mAP points more than frame-by-frame inference).
Conversely, our work introduces a novel probabilistic algorithm for aggregating scores over time, while these methods make the average and, differently from our solution, do not achieve any thorough improvement concerning the baseline. 

More recently, transformer architectures have also been widely adopted for VOD tasks. 
The TransVOD Lite~\cite{Trasvod} uses Deformable DETR object detector~\cite{zhu2021deformable} and scores a mAP50 of  \SI{83.7}{\nothing} on ImagenetVID with a SwinT transformer backbone~\cite{Swin_arc}. 
The latter features a total of \SI{45.9}{\mega\nothing} parameters and a computational cost of \SI{8.89}{\giga\nothing}MAC per frame, exceeding the resources of an MCU system. 
In the best configuration, TransVOD Lite achieves +\SI{3.5}{\nothing} mAP score vs. the object detector baseline by processing a batch of \SI{15}{\nothing} frames. 
This setting requires \SI{272}{\mega\byte} to store the intermediate features (15$\times$ memory increases compared to single frame execution).
If applied on individual frames, the method, instead, reduces the mAP by \SI{3.6}{\nothing} w.r.t. the  Deformable DETR detector. 
On the contrary, our method (i) does not store multiple features across frames, but only the bounding boxes (\SI{9.5}{\kilo\byte}) and (ii) leads to an mAP increases vs. the baseline when running frame-by-frame, i.e., it does not require a batch of images for the inference.  

YOLOV~\cite{2023yolov} uses a transformer on top of a YOLOX detector to fuse the features extracted from multiple frames and refine the produced detection. 
This approach reaches a mAP of \SI{85.5}{\nothing} on ImagenetVID. However, tuning the transformer requires a training stage, marking a significant difference w.r.t. our training-free method, which can be applied to any pre-trained model without additional computation.

\subsection{Increasing VOD Throughput}

A set of methods proposed data-dependent DNN execution pipelines to skip part of the inference operations based on the information carried by the input. The work of \textit{Zhu et al.}~\cite{DynamicRes} introduces an auxiliary but lightweight network, called dynamic-resolution network (DRNet), to determine the resolution of the input image before running the inference. 
The policy predicts a higher downsize ratio for images with ``easy" objects to recognize. 
Concerning this work, we adopt a static resize policy avoiding the \SI{4.9}{\mega\nothing} parameters overhead introduced by their DRNet, while achieving similar efficiency gains ($\sim$40\%).

On the contrary, other works speed up the processing by considering the temporal information and not only the content of a single frame.
\cite{Block_copy} avoids the computation of the features from specific regions of the current frame by copying the features computed in previous frames.
They use an auxiliary lightweight ResNet-8 network to predict the accuracy gain of computing new features vs. copying the previous ones. 
Compared to this approach, we do not store the activations but only preserve the detected bounding box coordinates from the analysis of previous ones. 
In the case of the NanoDet-Plus network with a $320\times320$ px input image, this method would lead to a memory overhead of $\sim$\SI{50}{\mega\byte} vs. the \SI{9.5}{\kilo\byte} of our method.
Furthermore, the ResNet-8 auxiliary network requires an extra training procedure using reinforcement learning to learn an efficient policy. 

\textit{Liu et al.}~\cite{Liu2019LookingFA} couple a large model with a small one (\SI{4.4}{\mega\nothing} vs. \SI{0.5}{\mega\nothing} parameters).
Similar to our approach, they also interleave the two models' execution, and to recover from accuracy losses due to the smaller model, they placed an LSTM layer to connect the detection performed across time. 
A similar idea is also employed in \cite{Motetti2024AdaptiveDL}, where they interleave different neural networks of varying complexity to determine the pose of a person in front of a drone.  
These methods deploy the two models on the target platforms, increasing the memory cost compared to the baseline single-model inference. On the contrary, our framework does not require any extra parameters. It is applied to a pre-trained model, saving the cost of training a family of scalable models for the interleaved execution.

%% file: 04-methodology.tex
\begin{figure*}[t]
\centering
\includegraphics[width=\textwidth]{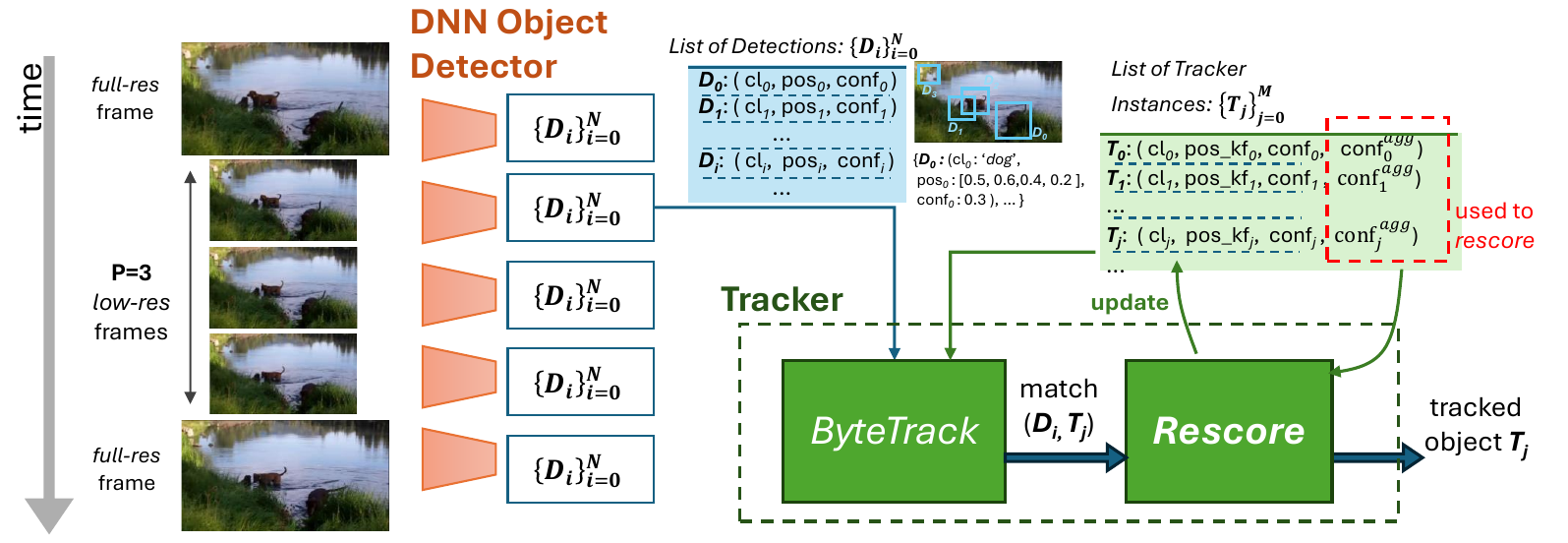}
\caption{Overview of the proposed Multi-Resolution Rescaled ByteTrack algorithm for video object detection. }
\label{fig:overview}
\end{figure*}

\section{Multi-resolution Rescored ByteTrack} \label{sec:system}

The MR2-ByteTrack algorithm in Fig.~\ref{fig:overview} combines three different components: (\textit{i}) a vision-based Convolutional Neural Network (CNN) object detector that takes in input images of different sizes; (\textit{ii}) the ByteTrack tracking algorithm~\cite{ByteTrackMT}; (\textit{iii}) the novel \textit{Rescore} algorithm to update the detection scores returned by ByteTrack.

\subsection{Multi-resolution Object Detector Backbone} \label{sec:method.objdec}

We design our framework around existing pre-trained single-shot object detectors. 
In our formulation, the detection algorithm includes a lightweight CNN model, i.e., the backbone, to analyze the image frames of a video sequence in real-time.
Given a frame $F$ at time $t$, we indicate the set of detections the processing task returns as $\{D_i\}_{i=1}^N$, where $N$ is the total number of objects detected within the current frame.
Each detection $D_i$ is described by a triplet $(\mathrm{bbox}_i, \mathrm{cl}_i, \mathrm{conf}_i)$: a bounding box $\mathrm{bbox}$, a class index $\mathrm{cl}$ and a confidence score  $\mathrm{conf}<1$.
Detections with a confidence score lower than a \textit{low\_threshold} are discarded, while the others are retained for further processing.

Since we employ a CNN-based backbone, we can interleave one full-res image (e.g., 320$\times$320 px in our setup) with $P$ low-res ones (e.g., 192$\times$192 px).
This way, we can trade detection accuracy for fewer MAC operations.
Thus, the average number of MAC operations per frame becomes: 

\begin{equation}\label{eq:mac}
\mathrm{MAC} = \rho \cdot \mathrm{MAC}^{fr} + (1-\rho)\cdot \mathrm{MAC}^{lr}
\end{equation}

Where $\rho = \frac{1}{1+P}$, $\mathrm{MAC}^{fr}$ and $\mathrm{MAC}^{lr}$ represent the number of MAC operations required to execute the inference on one full-res and one low-res frame, respectively.
Because the inference task \textit{utilizes the same network parameters} in both low and full-resolution cases, the weight footprint remains the same as with single-sized frames.
In fact, the convolution layers of the CNN models can apply to any scale of input images or feature maps. 
As an example, a 3x3 filter can slide over a larger feature map to produce a wider output tensor, corresponding, eventually, to a higher number of detections.

\subsection{ByteTrack Tracker} \label{sec:method.byte}

Na\"ive frame-by-frame object detectors can not exploit any temporal correlation between frames, i.e., detections in consecutive frames are likely to stay the same or similar.
We use the ByteTrack~\cite{ByteTrackMT} tracker to exploit this correlation, recovering for performance losses due to down-scaling (i.e., low-res images) and possibly improving the precision and recall of a frame-by-frame baseline (see Sec.~\ref{sec:multi-resol-results}).
The backbone's detections $\{D_i\}_{i=1}^N$ are fed to the ByteTrack algorithm based on a Kalman filter for every frame.
In its original formulation, at time-step $t$, ByteTrack operates on a list of $M$ instances: $\{T_j\}_{j=1}^M$, i.e., one instance for each tracked object.
Like for detections, each tracker instance ($T_j$) is described by a triplet  $(\mathrm{bbox\_kf}_j, \mathrm{cl}_j, \mathrm{conf}_j)$.
$\mathrm{bbox\_kf}_j$ is the bounding box predicted by a Kalman filter fed with the backbone's detections.
$\mathrm{cl}_j$ and $\mathrm{conf}_j$ are the class index and confidence scores associated with tracked objects, respectively. 

To correlate the detection performed over time, ByteTrack matches the $\mathrm{bbox}$ bounding boxes of $\{D_i\}_{i=1}^N$ and the predicted $\mathrm{bbox\_kf}$ ones of $\{T_j\}_{j=1}^M$, using a cost matrix determined by Intersection-over-Union (IoU) scores. 
An IoU threshold of 0.3, as proposed in the original paper, sets the minimum value to determine a match between a detection and a tracker instance.
Unlike the original implementation, we do not adopt any feature-matching criteria to associate tracker instances and detections: the spatial size of the feature maps to match changes over time in our multi-resolution framework. 
A tracker instance is marked active after a minimum number of consecutive detections has occurred (two in our setting). 
At the same time, it is removed from the tracker list if it was never updated during the last five time steps.
After matching, the detection $\mathrm{bbox}$ updates ByteTrack's Kalman filter.
The remaining detections not matched with any existing tracker instance produce a new one if their $\mathrm{conf}$ score is higher than a tuned \textit{high\_threshold}.

In the original ByteTrack algorithm, the class index $\mathrm{cl}_j$ and the confidence score $\mathrm{conf}_j$ of active tracker instances are never updated after the match with a new detection. 
We chose to modify this logic to better cope with the multi-class nature of VOD: specifically, we update $\mathrm{cl}_j$ and $\mathrm{conf}_j$ if the confidence score of the matched detection is higher than the \textit{high\_threshold} value.

\subsection{Rescore Algorithm}\label{sec:rescore}

\RestyleAlgo{ruled}
\begin{algorithm}[t]
\caption{Rescore algorithm}
\label{algo}
{\footnotesize
$\mathrm{\textbf{Inputs: }}$ Match $ \{ D_i = (\mathrm{cl}_i, \mathrm{conf}_i), T_j = (\mathrm{cl}_j, \mathrm{conf}_j, \mathrm{conf_j^{agg}}) \}$ \\
\eIf {$\mathrm{cl}_j$ == $\mathrm{cl_i}$ }
    {$\mathrm{conf_j^{agg}} \gets 1-((1-\mathrm{conf}_i)*(1-\mathrm{conf_j^{agg}}))$}
    {\eIf{$\mathrm{conf_j^{agg}} < \mathrm{conf}_i$}
    {
    $(\mathrm{cl}_j, \mathrm{conf}_j, \mathrm{conf^{agg}}_j ) \gets (\mathrm{cl}_i, \mathrm{conf}_i, \mathrm{conf}_i)$\\
    }
    {
    $ \mathrm{conf_j^{agg}} \gets 1-((1-\mathrm{conf_j^{agg}})/(1-\mathrm{conf}_i))$\\
    $\mathrm{conf_j^{agg}} \gets max(\mathrm{conf_j^{agg}},0)$ \\
    \If{$\mathrm{conf_j^{agg}} < \mathrm{conf}_i$}
        {
        $(\mathrm{cl}_j, \mathrm{conf}_j, \mathrm{conf_j^{agg}} ) \gets (\mathrm{cl}_i, \mathrm{conf}_i, \mathrm{conf}_i)$
}
        }
    }
        $\mathrm{conf_j^{agg}} \gets min(\mathrm{conf_j^{agg}},1-\epsilon)$\\
        \Return $T_j = (\mathrm{cl}_j, \mathrm{conf}_j, \mathrm{conf_j^{agg}}) $
}
\end{algorithm}

We augment ByteTrack with the \textit{Rescore} algorithm to improve the estimation of the class index and confidence scores of an active tracker instance by accounting for the history of matched detections. 
To this aim, we extend the tracker instance $T_j$ with a class status attribute $\mathrm{conf_j^{agg}}$, which models the probability of the tracker's class $\mathrm{cl}$ to be correct. 
The $\mathrm{conf_j^{agg}}$ value aggregates the confidence scores $\mathrm{conf}$ of the detections assigned to the $j$-th tracker instance until time $t-1$. 
During this aggregation, the $\mathrm{conf}$ value is interpreted as the probability that the object's predicted class is correct in the current-frame detection.

When a new detection $D_i$ is assigned to the tracker instance $T_j$
at time $t$, the class index $\mathrm{cl_j}$ is \textit{rescored} based on $\mathrm{conf_j^{agg}}$ as shown in Algo.~\ref{algo}.
If the class index $\mathrm{cl_i}$ is the same as $\mathrm{cl}_j$, we only update the $\mathrm{conf_j^{agg}}$ value, as the match confirms our confidence in the correctness of the classification.
We use the product of the opposites of $\mathrm{conf_j^{agg}}$ and $\mathrm{conf}_i$ as the new value for the status variable.
If the class indices do not match, we acknowledge this by decreasing our aggregate confidence, i.e., we decrease the $\mathrm{conf_j^{agg}}$ value (the $\mathrm{max}$ guarantees a value $>$0).
Then, we check for a class index rescore: if the new detection's confidence is higher than the $\mathrm{conf_j^{agg}}$, we update the tracker class with the new detection: $\mathrm{cl_j} \gets \mathrm{cl_i}$.
Additionally, we impose an upper bound (i.e., $1-\epsilon$) to the $\mathrm{conf_j^{agg}}$ to prevent overconfident detections from blocking future rescores. 
Finally, the \textit{Rescore} algorithm also outputs the mean confidence score during the tracking.

%% file: 05-GAP9.tex
\section{Multi-resolution Inference on MCUs}

We target ultra-low-power embedded systems, such as tiny MCUs, to deploy our pipeline. 
Since the end-to-end latency of VOD applications is dominated by DNN inference, we extensively benefit from the throughput optimization of Eq.~\ref{eq:mac}.
To demonstrate our approach's real-time performance and power consumption, we select a GWT GAP9 MCU, which includes a parallel cluster of ~9 general-purpose RISC-V cores. 
The cores can access 4 shared floating point units with vectorized half-precision instructions (\texttt{FP16} precision).
GAP9 features a low-latency scratchpad L1 memory of \SI{128}{\kilo\byte} within the cluster and a larger off-cluster L2 memory (\SI{1.5}{\mega\byte}).

A Direct Memory Access (DMA) controller is used to copy data between L2 and L1 memories in the background of the compute tasks offloaded to the cores. 
External volatile and non-volatile memories are interfaced with the GAP9 using the available octoSPI interface: a peripheral DMA is then in charge of read and write operations between the external components and the on-chip L2 memory.

The cluster also includes a convolution accelerator that supports 8-bit operations. 
However, we only exploit the floating-point multi-core support of the platform to avoid any accuracy degradation of lossy 8-bit quantization, which we leave for future work. 

We use GWT's GAP\textit{flow} toolset to generate the target platform's inference \texttt{C} code.
For every trained model with a specific input size, the tool produces header files with the model parameter and a source file with the graph-level and layer-level routines. 
Following the automatically generated memory management scheme, weight parameters are stored within an external octoSPI Flash memory. 
At the same time, the intermediate values of the inference tasks, i.e., the activation tensors, are allocated within the on-chip L2 memory and the external octoSPI RAM.
To cope with our multi-resolution scheme, we generate two \texttt{C} implementations, one for the low-res case and a second for the full-res one. 
The two network functions access the same weight data stored in the Flash memory.

On GAP9, the latency cost to downscale the input image is negligible with respect to the inference time. 
The overhead for generating a low-res image of 192$\times$192 px from a 320$\times$320 px image is  \SI{1.5}{\milli\second}. 
This corresponds to only 0.9\% of the inference time in the case of NanoDet-Plus (\SI{169}{\milli\second}).
The Kalman filter also has low execution time and requires only \SI{9.5}{\kilo\byte} of on-chip memory.
While our implementation of the MR2-ByteTrack method is specific to the GAP9, we remark on the generality of the method that can be easily applied to other MCUs.

%% file: 06-results.tex
\begin{table}[t]
\caption{Baseline Object Detection Models}
\label{tab:overview}
\resizebox{\columnwidth}{!}{
\begin{tabular}{l | c| cc| cc}
\toprule
\textbf{Model} & Params & full-res (px) & $\mathrm{MMAC}^{fr}$ & low-res (px) & $\mathrm{MMAC}^{lr}$ \\
\midrule
YOLOX-Nano & \SI{0.9}{\mega\nothing} & 320$\times$320 & 316 & 192$\times$192 & 114 \\
NanoDet–Plus & \SI{1.18}{\mega\nothing} & 320$\times$320 & 463 & 192$\times$192 & 167 \\
EfficientDet-D0 & \SI{3.93}{\mega\nothing} & 384$\times$384 & 1440 & 256$\times$256 & 640 \\
\bottomrule 
\end{tabular}}
\end{table}

\begin{table}[t]
\centering
\caption{Impact of the Rescore Algorithm on ByteTrack using only high-res frames ($P=0$).}
\label{tab:abls}
\resizebox{\columnwidth}{!}{
\begin{tabular}{l|c|ccc} 
\toprule
\textbf{Method}            & \textbf{mAP}  & \textbf{Precision} & \textbf{Recall} & \textbf{F1 score} \\ \midrule
YOLOX-Nano     & 41.30        &  \textbf{65.16}          & 44.92     & 53.18  \\
\textit{w/ Na\"ive-ByteTrack} & 43.57        &  63.9          & 47.96      & 54.79   \\
\textit{w/ MR2-ByteTrack} & \textbf{46.47}        &  64.24          & \textbf{51.23      }& \textbf{57.00}   \\
\hline 
NanoDet–Plus       & 42.70         &  62.15             & 47.68     & 54.00  \\
\textit{w/  Na\"ive-ByteTrack} & 44.00         &  58.91           & 51.40      & 54.90   \\
\textit{w/  MR2-ByteTrack}    & \textbf{46.57}       &  \textbf{64.14}           & \textbf{51.50}     & \textbf{57.10}    \\
\hline
EfficientDet-D0    & 60.70        &  78.48         & 64.14     & 70.30  \\
\textit{w/  Na\"ive-ByteTrack} & 61.09         &  78.65           & 66.42      & 72.02   \\
\textit{w/  MR2-ByteTrack} & \textbf{64.86}         &  \textbf{80.63}         & \textbf{67.32}      & \textbf{72.52} \\
\bottomrule
\end{tabular}}
\end{table}

\section{Results} \label{sec:results}

\begin{table*}[t]

\caption{VOD solutions on the GAP9 MCU: Mr2-ByteTrack vs. frame-by-frame CNN object detectors with full-res or low-res inputs. }
\label{tab:mcu_comp}

\resizebox{\textwidth}{!}{
\begin{tabular}{c|c|c|c|c|c|c|c|c|c|c}
\toprule
\textbf{Method}     &\textbf{\begin{tabular}[c]{@{}c@{}}Input\\ Size \end{tabular}}              & \textbf{\begin{tabular}[c]{@{}c@{}} RAM \\{[}\SI{}{\mega \byte}{]}\end{tabular}} & \textbf{ \begin{tabular}[c]{@{}c@{}} Flash \\Mem. {[}\SI{}{\mega \byte}{]}\end{tabular}} & \textbf{\begin{tabular}[c]{@{}c@{}} Code \\ {[}\SI{}{\kilo \byte}{]}\end{tabular}} & \textbf{{[}FPS{]}} & \textbf{\begin{tabular}[c]{@{}c@{}} Energy \\{[}\SI{}{\milli \joule}{]}\end{tabular}} & \textbf{mAP} & \textbf{Precision} & \textbf{Recall} & \textbf{\begin{tabular}[c]{@{}c@{}}F1\\ score\end{tabular}} \\ \midrule
\textit{NanoDet–Plus}      & \textit{single} full-res           & 1.6                   & 2.3                            & 186.0                         &3.3                              & 21.8                     & 42.7         & 62.2               & 47.7            & 54.0             \\
\textit{NanoDet–Plus}  & \textit{single} low-res           & 0.6                   & 2.3                            & 186.0                         & \textbf{9.8}                          & \textbf{7.5}                      & 27.4         & \textbf{62.9}               & 31.3            & 41.8       \\
 \textit{w/ MR2-ByteTrack} (our) & \textit{multi-res} P=2 & 1.6                   & 2.3                            & 373.0                         &5.9                          & 12.3                     & \textbf{44.9}        & 61.5               & \textbf{49.3}            & \textbf{54.7}            \\\hline
\textit{YOLOX-Nano} & \textit{single} full-res                     & 1.4                   & 1.8                            & 103.0                         & 3.3                            & 19.2                     & 41.3         & \textbf{65.2}              & 44.9            & 53.2          \\
\textit{YOLOX-Nano} & \textit{single} low-res              & 0.5                  & 1.8                            & 103.0                        &\textbf{8.6}                             & \textbf{7.0 }                       & 25.7         & 63.2               & 28.8            & 39.6        \\
\textit{w/ MR2-ByteTrack} (our) & \textit{multi-res} P=2& 1.4                   & 1.8                            & 205.0                         &5.5                           & 11.1                     & \textbf{41.4}        & 64.0                & \textbf{45.7}            & \textbf{53.3}              \\
\bottomrule
\end{tabular}}
\end{table*}

\subsection{Baseline Models and Metrics}

We consider three State-of-the-Art object detectors trained on the COCO dataset: YOLOX-Nano, NanoDet-Plus, and EfficientDet-D0.
Tab.~\ref{tab:overview} reports the number of parameters and \SI{}{\mega MAC} operations for the input frames' considered full-res and low-res sizes. 
For VOD experiments, we use the ImageNetVID dataset, which shares 16 classes with the COCO dataset.
By extracting video sequences that share common classes between the two datasets, we build our validation set of 392 video samples.
To distinguish from the entire dataset, we indicate this subset as ImageNetVID$^C$.
To assess the detection performance, we use the mAP50 score, a widely used metric in literature, which we call mAP in the rest of the paper.
In addition, we report the precision, recall, and F1 score averaged over the classes. 
Differently from mAP, which expresses the mean value of the per-class average-precision curves, these three additional metrics show the trade-off between true detections, false positives, and miss-detections. 
All metrics consider an IoU of 0.5 to mark a correct match between ground truths and predictions.

\subsection{Single-resolution Trackers}

Tab.~\ref{tab:abls} compares the results obtained with our MR2-ByteTrack against the baseline object detection models and the Na\"ive-ByteTrack on the ImageNetVid$^C$ dataset.
In this initial experiment, we feed the inference models only with full-res images.
The confidence threshold of the baseline models is set to 0.35, 0.3, and 0.4 for, respectively, NanoDet-Plus, YOLOX-Nano, and EfficientDet-D0. 
These values are optimally tuned to obtain the highest F1 score on the testing dataset and reflected on the \textit{high\_threshold} value of the ByteTrackers. 
Using the same procedure, we also set the \textit{low\_threshold} of the trackers to 0.3, 0.25, 0.35 for NanoDet-Plus, YOLOX-Nano, and EfficientDet-D0, respectively. By observing the results, we can see that, on average,  the Na\"ive-ByteTrack increases the mAP score measured on the baseline models by \SI{1.55}{\nothing}\%.
This is explained by the capacity of the tracker to positively aggregate low-confident detections, which the baseline models instead discard. 

Next, the Rescore algorithm of MR2-ByteTrack leads to a further 3.08\% improvement by turning false-positive detection into correct classification, as can be observed by the higher Precision, Recall, and F1 scores. 
Overall, thanks to our MR2-ByteTrack, the smallest NanoDet-Plus and YOLOX-Nano models can achieve up to \SI{46.57}{\nothing} and \SI{46.47}{\nothing} mAP on high-res frames, respectively, +\SI{3.9}{\nothing}\% and +\SI{5.2}{\nothing}\% vs. the baselines.

\begin{table*}[t]
\centering
\caption{Comparison between VOD methods on ImagenetNetVID.}
\label{tab:comp}
\resizebox{1\textwidth}{!}{
\begin{threeparttable}
\begin{tabular}{l|c|c|c|c|c|c|c|c|c|c}
\toprule
\multirow{2}{*}{\textbf{Method}} & \multirow{2}{*}{\textbf{\begin{tabular}[c]{@{}c@{}} Object detector\\ Backbone\end{tabular}}} & \multirow{2}{*}{\textbf{Prec.}} & \multirow{2}{*}{\textbf{Recall}} & \multirow{2}{*}{\textbf{\begin{tabular}[c]{@{}c@{}}F1\\ score\end{tabular}}} & \multirow{2}{*}{\textbf{mAP}} & \multirow{2}{*}{\textbf{\begin{tabular}[c]{@{}c@{}}Params\\ {[}M{]}\end{tabular}}} & \multirow{2}{*}{\textbf{GMAC}} & \multicolumn{3}{c}{\textbf{vs. full-res object detetector}}                                      \\ \cline{9-11} 
                                 &                                                                                       &                                 &                                  &                                                                              &                               &                                                                                    &                                & \multicolumn{1}{c|}{$\Delta$ mAP} & \multicolumn{1}{c|}{$\Delta$ params} & $\Delta$ MAC \\ \hline
YOLOV-S    \cite{2023yolov}                           &                     & 79.9               & 66.4            & 72.5              & 62.5          & 9.9                   & 12.9          &2.7                  &  +14.0\% & +21.0\%                  \\
\begin{tabular}[c]{@{}c@{}} MR2-ByteTrack \end{tabular}            &                  & 80.8               & 65.9            & 72.5              & 61.8        & 9.0                    & 10.9          & 2.0                   & 0                      & 0                  \\
\begin{tabular}[c]{@{}c@{}} MR2-ByteTrack (P=2)\end{tabular}         & \multirow{-3}{*}{YOLOX-S\cite{ge2021yolox} }                 & 77.5               & 64.3            & 70.3              & 60.3        & 9.0                    & 6.2          & 0.5                 & 0                      & -43.0\%              \\ \midrule
\textit{Liu et. al}  \cite{Liu2019LookingFA}                          & SSDLite-Mobilenetv2\cite{MobileNetv2}       & n/a                & n/a             & n/a               & 61.4         & 4.9                     & 0.2          & 0.9$^1$                 & +11.0 \% $^1$                 & -84.0\%  \\ 
\begin{tabular}[c]{@{}c@{}}MR2-ByteTrack (P=2) \end{tabular}                   & EfficientDet-D0$^3$~\cite{EfficientDET}        &  80.0         & 64.1    & 71.1                & 61.2$^2$         & 3.9                     & 0.9         & 0.5$^2$                 & 0                 & -37.0\%  \\ 
\bottomrule
\end{tabular}
\begin{tablenotes}
{
    \item[1] w.r.t. large f$_0$ non-interleaved model ~\cite{Liu2019LookingFA},
    \item[2] measured on ImageNetVID$^C$,
    \item[3] trained on the COCO dataset,
    }
\end{tablenotes}
\end{threeparttable}}
\end{table*}

\subsection{Multi-resolution Performance} \label{sec:multi-resol-results}

\begin{figure}[t]
  \includegraphics[width=\columnwidth]{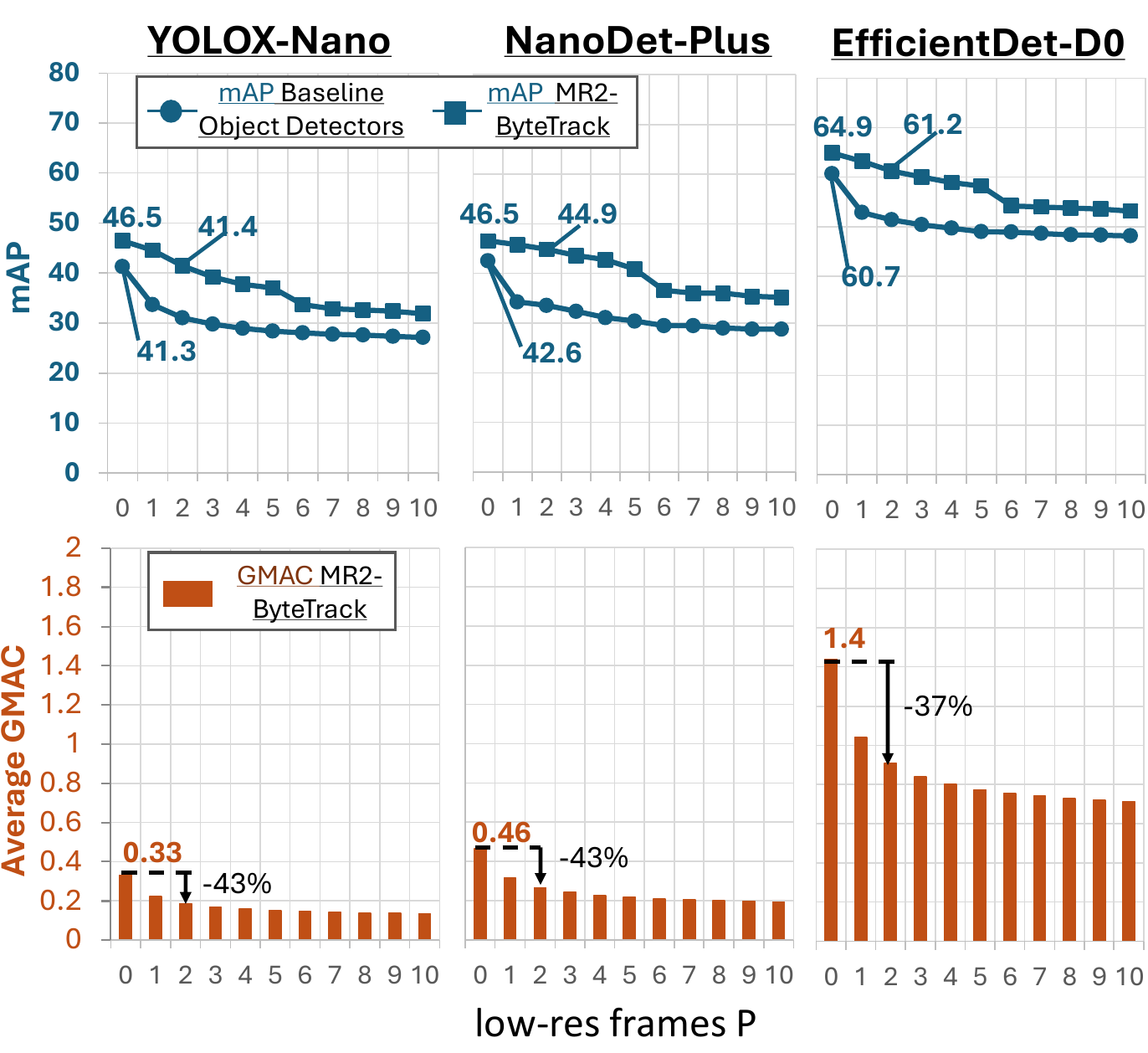}
  \caption{mAP (blue) and GMAC (red) of MR2-ByteTrack at varying low-res frames vs. the baseline.}
  \label{fig:result_resize}
\end{figure}

Fig.~\ref{fig:result_resize} plots the measured mAP of MR2-ByteTrack when varying the amount of interleaved low-res images ($P$ from 0 to 10) in the multi-resolution setup, and we benchmark against the baseline models in the same setup. 
The bars on the bottom show the average \SI{}{\giga MAC} operations required by each configuration (Eq.~\ref{eq:mac}).
For baseline models, the mAP starts dropping at $P$=1, while MR2-ByteTrack decreases performance with $P\ge$ 5 due to low-res images.
This difference between methods arises from the greater robustness of our MR2-ByteTrack approach.
This effect is related to the inertia of the trackers, which are kept alive for up to 5 timestamps without new detections in our setup.

Overall, the mAP scores of the MR2-ByteTrack models using multiple low-res images are always better than the baseline models operating on full-resolution frames. 
For $P$=1 and $P$=2, on average, our approach marks a mAP improvement of \SI{3}{\nothing}\% and \SI{0.96}{\nothing}\%, while the number of MAC operations is reduced by \SI{32}{\nothing}\% and \SI{43}{\nothing}\%, respectively, compared with the baseline object detectors.
For $P\geq$ 3, we observe a progressive reduction of mAP below the original object detection performance, which can be affordable for some use cases. 
Finally, we chose the configuration $P$=2 for the remainder of our experiments as a good trade-off between accuracy and computational load (i.e., MAC operations).

\subsection{State of the Art Comparison }

\subsubsection{MCU-ready VOD Systems}

Tab.~\ref{tab:mcu_comp} compares our MR2-ByteTrack solution with $P$=2 vs. frame-by-frame single-resolution baselines (e.g., used by \cite{lamberti2021low,lamberti2023bio}) with full-res and low-res inputs when running on an MCU system.
To this aim, we deploy the different approaches on the GAP9 SoC, coupled with two external memories: a Flash of \SI{64}{\mega\byte} and a RAM of \SI{8}{\mega\byte}. 
The table reports the code size, the activation and weight memory occupation in MB, denoted as RAM and FLASH memory, the processing throughput in FPS, and the energy cost, measured when the MCU is clocked at \SI{370}{\mega\hertz}, together with the detection metrics.

We show results for YOLOX-Nano and NanoDet-Plus, while we leave out of this analysis the EfficientDet, as its memory footprint is larger than the available budget.
The datatype of weights and activations is cast to half-precision \texttt{FP16} for a lossless deployment.
On GAP9, our multi-resolution instance features a double code size (up to \SI{373}{\kilo\byte} in total) vs. baseline models.
The activation memory of the high-res model dominates the total cost of the multi-res deployment; the weight storage is constant among different resolutions. 
Our solutions can run at \SI{5.9}{FPS} with NanoDet-Plus and \SI{5.5}{FPS} with YOLOX-Nano, 1.7$\times$ faster than the high-res version. 
On the other side, the cost of the ByteTrack algorithm is negligible ($<$0.2\% than the inference). 
The latency gains are reflected in the energy costs: the multi-resolution models consume 1.77$\times$ and 1.72$\times$ less than the high-res versions for, respectively, NanoDet-Plus and  YOLOX-Nano. 
Thus, observing the highest mAP and F1 scores, \textit{our MR2-ByteTraker shows the best energy-accuracy trade-off for VOD applications on MCU systems}.

\subsubsection{VOD Algorithms} 

Tab.~\ref{tab:comp} compares our approach with recent real-time methods \textit{not} tailored for MCU deployment given the high number of parameters of the DNN backbones: \SI{9}{\mega\nothing} for YOLOX-S in the YOLOV work~\cite{2023yolov} and \SI{4.9}{\mega\nothing} for SSDLite-MobileNetV2 in the study by \textit{Liu et al.}~\cite{Liu2019LookingFA}. To fairly compare vs. YOLOV, which uses a transformer to aggregate the temporal detections of the YOLOX-S backbone, we apply our MR2-ByteTrack to the same CNN object detector pre-trained on the ImageNetVid trainset, considering both the high-res or the multi-resolution (P=2) scenarios. 
For YOLOV,~we test the real-time version in which only the detections of the past frames are sent to the transformers.
The scores reported in the table adopt confidence threshold values that maximize the F1 score. 

Our MR2-ByteTrack reaches a +0.9 higher precision than the YOLOV due to the lower number of false positives. 
At the same time, the lower recall (-0.5) acknowledges the recognition ability of the transformer layer, which increases the memory and computational cost by, respectively, 14\% and 21\% compared to the baseline backbones.
Our method presents the same F1 score as the real-time YOLOV, saving the transformer overhead costs.
Additionally, with P=2, the F1 score is reduced only by 2.2\% with a computational cost reduction of up to 43\%.

On the other side, because the pre-trained SSDLite-Mobilenetv2 model adopted by \textit{Liu et al.}~\cite{Liu2019LookingFA} is not openly available, we take the EfficientDet-D0 model pre-trained on COCO for comparison purposes (EfficientDet-D0 features \SI{1}{\mega\nothing} parameters less than SSDLite-Mobilenetv2). 
Our MR2-ByteTrack method achieves a substantial latency improvement (37\% 
vs. 84\% of~\cite{Liu2019LookingFA} that uses $P$=10) while slightly increasing the mAP score compared to the baseline.
However, our solution does not increase the memory footprint (+11\% in~\cite{Liu2019LookingFA}) and does not require any training of the tracker on the video sequence, highlighting the higher flexibility of our approach.

%% file: 07-conclusions.tex
\section{Conclusions} \label{sec:conclusions}

This paper proposes the novel MR2-ByteTrack method that (\textit{i}) improves the accuracy of pre-trained object detection for VOD benchmarks by using the ByteTrack tracker augmented with the novel rescore algorithm and (\textit{ii}) reduces the computational cost by running multi-resolution inference at no extra memory costs. 
Leveraging SoA memory-efficient DNNs, we achieved equal or slightly increased mAP and up to 1.76$\times$ improvement in throughput on the GAP9 MCU, compared to full resolution single frame object detectors.
This result highlights the benefits of the proposed solution for VOD in ultra-low-power embedded devices.

%% file: main.bbl
\begin{thebibliography}{32}
\providecommand{\natexlab}[1]{#1}
\providecommand{\url}[1]{\texttt{#1}}
\expandafter\ifx\csname urlstyle\endcsname\relax
  \providecommand{\doi}[1]{doi: #1}\else
  \providecommand{\doi}{doi: \begingroup \urlstyle{rm}\Url}\fi

\bibitem[AlNuaimi et~al.(2022)AlNuaimi, Cereda, Psiakis, Sugumar, Giusti, and Palossi]{alnuaimi2022deep}
Eiman AlNuaimi, Elia Cereda, Rafail Psiakis, Suresh Sugumar, Alessandro Giusti, and Daniele Palossi.
\newblock A deep learning-based face mask detector for autonomous nano-drones (student abstract).
\newblock In \emph{Proceedings of the AAAI Conference on Artificial Intelligence}, pages 12903--12904, 2022.

\bibitem[Belhassen et~al.(2019)Belhassen, Zhang, Fresse, and Bourennane]{Seq-BBox}
Hatem Belhassen, Heng Zhang, Virginie Fresse, and El-Bay Bourennane.
\newblock Improving video object detection by seq-bboxmatching.
\newblock In \emph{VISIGRAPP(5:VISAPP)}, pages 226--233, 2019.

\bibitem[Chen et~al.(2020)Chen, Cao, Hu, and Wang]{MEGA}
Yihong Chen, Yue Cao, Han Hu, and Liwei Wang.
\newblock Memory enhanced global-local aggregation for video object detection.
\newblock In \emph{2020 IEEE/CVF Conference on Computer Vision and Pattern Recognition (CVPR)}, pages 10334--10343, 2020.

\bibitem[Ge et~al.(2021)Ge, Liu, Wang, Li, and Sun]{ge2021yolox}
Zheng Ge, Songtao Liu, Feng Wang, Zeming Li, and Jian Sun.
\newblock Yolox: Exceeding yolo series in 2021.
\newblock \emph{arXivpreprintarXiv:2107.08430}, 2021.

\bibitem[Han et~al.(2016)Han, Khorrami, Paine, Ramachandran, Babaeizadeh, Shi, Li, Yan, and Huang]{seqnms}
Wei Han, Pooya Khorrami, Tom~Le Paine, Prajit Ramachandran, Mohammad Babaeizadeh, Honghui Shi, Jianan Li, Shuicheng Yan, and Thomas~S. Huang.
\newblock Seq-nms for video object detection.
\newblock \emph{CoRR}, abs/1602.08465, 2016.

\bibitem[He et~al.(2021)He, Zhou, Li, Niu, Cheng, Li, Liu, Tong, Ma, and Zhang]{Trasvod}
Lu He, Qianyu Zhou, Xiangtai Li, Li Niu, Guangliang Cheng, Xiao Li, Wenxuan Liu, Yunhai Tong, Lizhuang Ma, and Liqing Zhang.
\newblock End-to-end video object detection with spatial-temporal transformers.
\newblock In \emph{Proceedings of the 29th ACM International Conference on Multimedia}, page 1507–1516, New York, NY, USA, 2021. Association for Computing Machinery.

\bibitem[Lamberti et~al.(2021)Lamberti, Rusci, Fariselli, Paci, and Benini]{lamberti2021low}
Lorenzo Lamberti, Manuele Rusci, Marco Fariselli, Francesco Paci, and Luca Benini.
\newblock Low-power license plate detection and recognition on a risc-v multi-core mcu-based vision system.
\newblock In \emph{2021 IEEE International Symposium on Circuits and Systems (ISCAS)}, pages 1--5. IEEE, 2021.

\bibitem[Lamberti et~al.(2023)Lamberti, Bompani, Kartsch, Rusci, Palossi, and Benini]{lamberti2023bio}
Lorenzo Lamberti, Luca Bompani, Victor~Javier Kartsch, Manuele Rusci, Daniele Palossi, and Luca Benini.
\newblock Bio-inspired autonomous exploration policies with cnn-based object detection on nano-drones.
\newblock In \emph{2023 Design, Automation \& Testin Europe Conference \& Exhibition (DATE)}, pages 1--6. IEEE, 2023.

\bibitem[Li et~al.(2021)Li, Li, Bai, Ren, Meng, and Yang]{motion_based}
Min Li, Linghan Li, Ruwen Bai, Junxing Ren, Bo Meng, and Yang Yang.
\newblock A motion-based seq-bbox matching method for video object detection.
\newblock In \emph{2021 IEEE Symposium on Computers and Communications (ISCC)}, pages 1--7, 2021.

\bibitem[Lin et~al.(2014)Lin, Maire, Belongie, Hays, Perona, Ramanan, Doll{\'a}r, and Zitnick]{cocodataset}
Tsung-Yi Lin, Michael Maire, Serge Belongie, James Hays, Pietro Perona, Deva Ramanan, Piotr Doll{\'a}r, and C.~Lawrence Zitnick.
\newblock Microsoft coco: Common objects in context.
\newblock In \emph{Computer Vision -- ECCV 2014}, pages 740--755, Cham, 2014. Springer International Publishing.

\bibitem[Liu et~al.(2019)Liu, Zhu, White, Li, and Kalenichenko]{Liu2019LookingFA}
Mason Liu, Menglong Zhu, Marie White, Yinxiao Li, and Dmitry Kalenichenko.
\newblock Looking fast and slow: Memory-guided mobile video object detection.
\newblock \emph{arXiv preprint arXiv:1903.10172}, 2019.

\bibitem[Liu et~al.(2021)Liu, Lin, Cao, Hu, Wei, Zhang, Lin, and Guo]{Swin_arc}
Z. Liu, Y. Lin, Y. Cao, H. Hu, Y. Wei, Z. Zhang, S. Lin, and B. Guo.
\newblock Swin transformer: Hierarchical vision transformer using shifted windows.
\newblock In \emph{2021 IEEE/CVF International Conference on Computer Vision (ICCV)}, pages 9992--10002, Los Alamitos, CA, USA, 2021. IEEE Computer Society.

\bibitem[Lyu et~al.(2021)Lyu, Yang, Vosselman, and Xia]{LYU2021139}
Ye Lyu, Michael~Ying Yang, George Vosselman, and Gui-Song Xia.
\newblock Video object detection with a convolutional regression tracker.
\newblock \emph{ISPRS Journal of Photogrammetry and Remote Sensing}, 176:\penalty0 139--150, 2021.

\bibitem[Motetti et~al.(2024)Motetti, Crupi, Elshaigi, Risso, Pagliari, Palossi, and Burrello]{Motetti2024AdaptiveDL}
Beatrice~Alessandra Motetti, Luca Crupi, Mustafa Omer Mohammed~Elamin Elshaigi, Matteo Risso, Daniele~Jahier Pagliari, Daniele Palossi, and Alessio Burrello.
\newblock Adaptive deep learning for efficient visual pose estimation aboard ultra-low-power nano-drones.
\newblock \emph{ArXiv}, abs/2401.15236, 2024.

\bibitem[Portilla et~al.(2019)Portilla, Mujica, Lee, and Riesgo]{Extreme_edge}
Jorge Portilla, Gabriel Mujica, Jin-Shyan Lee, and Teresa Riesgo.
\newblock The extreme edge at the bottom of the internet of things: A review.
\newblock \emph{IEEE Sensors Journal}, PP:\penalty0 1--1, 2019.

\bibitem[RangiLyu(2021)]{nanodet}
RangiLyu.
\newblock Nanodet-plus superfast and high accuracy lightweight anchor-free object detection model.
\newblock 2021.

\bibitem[Redmon and Farhadi(2018)]{redmon2018yolov3}
Joseph Redmon and Ali Farhadi.
\newblock Yolov3: An incremental improvement, 2018.
\newblock cite arxiv:1804.02767Comment: Tech Report.

\bibitem[Ren et~al.(2017)Ren, He, Girshick, and Sun]{renNIPS15fasterrcnn}
S. Ren, K. He, R. Girshick, and J. Sun.
\newblock Faster r-cnn: Towards real-time object detection with region proposal networks.
\newblock pages 1137--1149, Los Alamitos, CA, USA, 2017. IEEE Computer Society.

\bibitem[Rossi et~al.(2022)Rossi, Conti, Eggiman, Mauro, Tagliavini, Mach, Guermandi, Pullini, Loi, Chen, Flamand, and Benini]{VegaDSP}
Davide Rossi, Francesco Conti, Manuel Eggiman, Alfio~Di Mauro, Giuseppe Tagliavini, Stefan Mach, Marco Guermandi, Antonio Pullini, Igor Loi, Jie Chen, Eric Flamand, and Luca Benini.
\newblock Vega: A ten-core soc for iot endnodes with dnn acceleration and cognitive wake-up from mram-based state-retentive sleep mode.
\newblock \emph{IEEE Journal of Solid-State Circuits}, 57\penalty0 (1):\penalty0 127--139, 2022.

\bibitem[Russakovsky et~al.(2015)Russakovsky, Deng, Su, Krause, Satheesh, Ma, Huang, Karpathy, Khosla, Bernstein, Berg, and Fei-Fei]{ILSVRC15}
Olga Russakovsky, Jia Deng, Hao Su, Jonathan Krause, Sanjeev Satheesh, Sean Ma, Zhiheng Huang, Andrej Karpathy, Aditya Khosla, Michael Bernstein, Alexander~C. Berg, and Li Fei-Fei.
\newblock {ImageNet Large Scale Visual Recognition Challenge}.
\newblock \emph{International Journal of Computer Vision (IJCV)}, 115\penalty0 (3):\penalty0 211--252, 2015.

\bibitem[Sandler et~al.(2018)Sandler, Howard, Zhu, Zhmoginov, and Chen]{MobileNetv2}
Mark Sandler, Andrew~G. Howard, Menglong Zhu, Andrey Zhmoginov, and Liang-Chieh Chen.
\newblock Mobilenetv2: Inverted residuals and linear bottlenecks.
\newblock \emph{2018 IEEE/CVF Conference on Computer Vision and Pattern Recognition}, pages 4510--4520, 2018.

\bibitem[Shi et~al.(2023)Shi, Wang, and Guo]{2023yolov}
Yuheng Shi, Naiyan Wang, and Xiaojie Guo.
\newblock Yolov: Making still image object detectors great at video object detection.
\newblock \emph{Proceedings of the AAAI Conference on Artificial Intelligence}, 37\penalty0 (2):\penalty0 2254--2262, 2023.

\bibitem[Sola-Thomas and Imtiaz(2021)]{ULP_Camera}
Ernesto Sola-Thomas and Masudul~Haider Imtiaz.
\newblock An ultra-low-power design of smart wearable stereo camera.
\newblock In \emph{SoutheastCon 2021}, pages 1--8, 2021.

\bibitem[Tan et~al.(2020)Tan, Pang, and Le]{EfficientDET}
M. Tan, R. Pang, and Q.~V. Le.
\newblock Efficientdet: Scalable and efficient object detection.
\newblock In \emph{2020 IEEE/CVF Conference on Computer Vision and Pattern Recognition (CVPR)}, pages 10778--10787, Los Alamitos, CA, USA, 2020. IEEE Computer Society.

\bibitem[Tran et~al.(2015)Tran, Bourdev, Fergus, Torresani, and Paluri]{tran2015learning}
D. Tran, L. Bourdev, R. Fergus, L. Torresani, and M. Paluri.
\newblock Learning spatiotemporal features with 3d convolutional networks.
\newblock In \emph{2015 IEEE International Conference on Computer Vision (ICCV)}, pages 4489--4497, Los Alamitos, CA, USA, 2015. IEEE Computer Society.

\bibitem[Verelst and Tuytelaars(2021)]{Block_copy}
Thomas Verelst and Tinne Tuytelaars.
\newblock Blockcopy: High-resolution video processing with block-sparse feature propagation and online policies.
\newblock In \emph{2021 IEEE/CVF International Conference on Computer Vision (ICCV)}, pages 5138--5147, 2021.

\bibitem[Xiong et~al.(2021)Xiong, Liu, Gupta, Akin, Bender, Wang, Kindermans, Tan, Singh, and Chen]{Xiong_2021_CVPR}
Yunyang Xiong, Hanxiao Liu, Suyog Gupta, Berkin Akin, Gabriel Bender, Yongzhe Wang, Pieter-Jan Kindermans, Mingxing Tan, Vikas Singh, and Bo Chen.
\newblock Mobiledets:searching for object detection architectures for mobile accelerators.
\newblock In \emph{Proceedings of the IEEE/CVF Conference on Computer Vision and Pattern Recognition (CVPR)}, pages 3825--3834, 2021.

\bibitem[Zhang et~al.(2021)Zhang, Sun, Jiang, Yu, Yuan, Luo, Liu, and Wang]{ByteTrackMT}
Yifu Zhang, Pei Sun, Yi Jiang, Dongdong Yu, Zehuan Yuan, Ping Luo, Wenyu Liu, and Xinggang Wang.
\newblock Bytetrack: Multi-object tracking by associating every detection box.
\newblock In \emph{European Conference on Computer Vision}, 2021.

\bibitem[Zhang et~al.(2018)Zhang, Cheng, Lin, and Dai]{IntegratedOD}
Zheng Zhang, Dazhi Cheng, Xizhou Zhuand~Stephen Lin, and Jifeng Dai.
\newblock Integrated object detection and tracking with tracklet-conditioned detection.
\newblock \emph{ArXiv}, abs/1811.11167, 2018.

\bibitem[Zhu et~al.(2021{\natexlab{a}})Zhu, Han, Wu, Zhang, Nie, Lan, and Wang]{DynamicRes}
Mingjian Zhu, Kai Han, Enhua Wu, Qiulin Zhang, Ying Nie, Zhenzhong Lan, and Yunhe Wang.
\newblock Dynamic resolution network.
\newblock In \emph{Neural Information Processing Systems}, 2021{\natexlab{a}}.

\bibitem[Zhu et~al.(2017)Zhu, Wang, Dai, Yuan, and Wei]{FGFA}
Xizhou Zhu, Yujie Wang, Jifeng Dai, Lu Yuan, and Yichen Wei.
\newblock Flow-guided feature aggregation for video object detection.
\newblock In \emph{2017 IEEE International Conference on Computer Vision (ICCV)}, pages 408--417, 2017.

\bibitem[Zhu et~al.(2021{\natexlab{b}})Zhu, Su, Lu, Li, Wang, and Dai]{zhu2021deformable}
Xizhou Zhu, Weijie Su, Lewei Lu, Bin Li, Xiaogang Wang, and Jifeng Dai.
\newblock Deformable {\{}detr{\}}: Deformable transformers for end-to-end object detection.
\newblock In \emph{International Conference on Learning Representations}, 2021{\natexlab{b}}.

\end{thebibliography}
